# Case-based reasoning approach for diagnostic screening of children with developmental delays

Zichen Song, Jiakang Li, Songning Lai, Sitan Huang
(Co-First Author)

**Abstract.** According to the World Health Organization, the population of children with developmental delays constitutes approximately 6% to 9% of the total population. Based on the number of newborns in Huaibei, Anhui Province, China, in 2023 (94,420), it is estimated that there are about 7,500 cases (suspected cases of developmental delays) of suspicious cases annually. Early identification and appropriate early intervention for these children can significantly reduce the wastage of medical resources and societal costs. International research indicates that the optimal period for intervention in children with developmental delays is before the age of six, with the golden treatment period being before three and a half years of age. Studies have shown that children with developmental delays who receive early intervention exhibit significant improvement in symptoms; some may even fully recover. This research adopts a hybrid model combining a CNN-Transformer model with Case-Based Reasoning (CBR) to enhance the screening efficiency for children with developmental delays. The CNN-Transformer model is an excellent model for image feature extraction and recognition, effectively identifying features in bone age images to determine bone age. CBR is a technique for solving problems based on similar cases; it solves current problems based on past experiences, similar to how humans solve problems through learning from experience. Given CBR's memory capability to judge and compare new cases based on previously stored old cases, it is suitable for application in support systems with latent and variable characteristics. Therefore, this study utilizes the CNN-Transformer-CBR to establish a screening system for children with developmental delays, aiming to improve screening efficiency. System validation showed that the case similarity reasoning average was 0.92, and the accuracy reasoning average was 0.91, indicating that the system's validation results are of a high level, thereby verifying the system's high feasibility.

---

[1] Email: 3528630668@qq.com

# 1 Introduction

In the process of children's growth and development, a multitude of factors interact to shape their physical, intellectual, and emotional well-being. Notably, during pregnancy, genetic factors along with chemical and physical environmental changes can impact the fetus, sometimes leading to challenges in social adaptation and learning patterns in children. Additionally, factors such as the childbirth process, postnatal nurturing practices, and the social environment in which they grow up play significant roles in child development. These early influences not only affect children's immediate learning and growth but can also have profound impacts on their future personality development, socio-economic status, and relationships with others. Thus, these developmental obstacles pose extra challenges and pressures for parents and families, as well as significant burdens for society and the nation at large.

In medical terms, "growth" typically refers to increases in height, weight, and organ volume, while "development" focuses on changes, improvements, and maturation in organ function, intelligence, and various skills. If children do not acquire age-appropriate skills during critical stages of growth and development, it can adversely affect their subsequent growth. Therefore, infancy is considered the most malleable phase of life. Early diagnosis and timely intervention can help children overcome developmental delays as soon as possible, especially in the first year after birth, when children are highly adaptable in cognitive development and learning capabilities. Timely early intervention not only helps prevent the worsening of issues but can also reduce the severity of disabilities, thereby avoiding adverse effects on future development and learning.

According to data from the World Health Organization (WHO), developmental delays account for a certain proportion of children under six. In Taiwan, this proportion is relatively low, possibly because current statistics mainly reflect infants and toddlers already identified as having disabilities, overlooking those in high-risk groups who have not been formally diagnosed. Moreover, due to the imperfection of early screening and diagnostic systems, many children in need of assistance do not receive timely intervention. From a long-term developmental perspective, developmental delays are not irreversible; children still have the potential for recovery and improvement. Appropriate early intervention programs can effectively prevent children from becoming individuals with significant intellectual delays. Given this, the government has been vigorously promoting early intervention programs in recent years, aiming to reduce the burden of medical and social costs. Despite the support of laws and policies, the diagnosis, treatment, and education of preschool children with developmental delays still face many challenges in practice. Issues such as vague definitions of government policies, unclear responsibilities of implementing agencies, and a severe lack of professional assessment and treatment services significantly affect the quality and coverage of early intervention measures.

This study utilizes a CNN-Transformer-CBR database system for screening new cases and enhances the efficiency of screening and diagnosis through early intervention centers. By integrating processes of early detection, screening assessment, report submission, and early intervention with continuous follow-up, this study aims to improve the efficiency of identifying and diagnosing children with developmental delays, reduce the wastage of medical resources, and enhance the quality of care, thereby providing a stronger foundation for the healthy growth of children.

## 2      Related Work

### 2.1    Childhood developmental delays and Case-Based Reasoning (CBR)

Recent research on developmental delays in children indicates that such delays can affect multiple areas of development, including but not limited to organ function, sensation, motor balance, language communication, cognitive learning, sociopsychology, emotions, and intelligence. It has been noted that when a child's developmental delay exceeds 20% of their chronological age, further professional evaluation is warranted, with studies from Taiwan serving as a reference. Although Taiwan primarily classifies children with developmental delays based on types of disabilities, this classification does not fully capture the extent of developmental delays. Therefore, it has been proposed that early screening and appropriate personalized treatments, or providing primary caregivers with advice on nurturing and training, could serve as primary or secondary preventive measures for children with developmental delays.

Early intervention aims to provide necessary support for children with developmental delays or disorders by integrating professional services to address their medical, educational, family, and social issues. The effectiveness of early intervention depends on the content of the treatment, the timing of the intervention, and the individual differences of each child. Studies have shown that early intervention can not only reduce life obstacles and psychological impacts on children, alleviate burdens on families and society, but also enhance individual functionality, independence, and life satisfaction, thereby effectively strengthening the nation's overall strength.

Moreover, Case-Based Reasoning (CBR) serves as a problem-solving technique by referring to past experiences to address current problems. The CBR system, by collecting and organizing cases in a database, provides a basis for solving new cases. As a knowledge-based system, CBR forms a valuable knowledge repository by collecting and processing experiences in the form of cases.

In the field of child development, screening scales are categorized based on concepts of developmental psychology, covering behaviors in language, social interaction, motor skills, and cognitive development from birth to six years of age. The primary purpose of these scales is to identify children who may have developmental difficulties, facilitating further diagnostic assessment and necessary follow-up.

**2.2 Bone Age Assessment**

Before the advent of deep learning, bone age assessment primarily relied on traditional image processing techniques. The Greulich and Pyle (G&P) method was one of the earliest and most widely used methods, evaluating bone age by comparing with standard reference image [1]. Additionally, the Tanner-Whitehouse (T-W) method offered a more detailed assessment of skeletal maturity through a scoring system [2]. While effective, these methods required significant manual intervention and were highly subjective in their results.

With the development of machine learning technology, researchers began exploring the automation of bone age assessment using feature extraction and pattern recognition techniques. For instance, algorithms such as Support Vector Machines (SVM) and Random Forests (RF) were used for classifying and regressing extracted features from hand X-ray images to predict bone age [3]. These approaches represented an improvement over traditional methods but were still limited by the efficiency and accuracy of manual feature extraction.

In recent years, deep learning-based methods have become a focal point of research in bone age assessment, particularly due to the remarkable performance of Convolutional Neural Networks (CNN) in image recognition tasks. These methods can automatically learn complex feature representations directly from raw X-ray images, significantly enhancing assessment accuracy and efficiency. For example, Lee and colleagues proposed an automatic bone age assessment system based on deep CNNs that directly predicts bone age from hand X-ray images, achieving significantly higher accuracy than traditional methods [4]. Moreover, research has shown that incorporating advanced features such as attention mechanisms can further improve model performance [5].

## 3   Method

**3.1   CNN-Transformer**

In this study, we utilized patient cases from the pediatric department of Huaibei People's Hospital in 2023 (1342 Images), with privacy measures implemented to prevent the disclosure of patient personal information. Additionally, we performed data preprocessing on these cases. During the data preprocessing phase, it was imperative to ensure the collected children's hand X-ray images encompassed a broad age range and diverse demographic characteristics to guarantee the model's generalizability and fairness. These images underwent thorough screening and quality control to ensure each picture had high resolution and clear skeletal structure for subsequent processing and analysis. [6]

The specific steps for image preprocessing included:
- **Resizing**: Standardizing the size of all images to fit the model input standard, selecting a dimension (224x224 pixels) that balances computational efficiency and detail preservation.

**- Normalizing brightness and contrast**: Adjusting the brightness and contrast of each image to reduce variations due to different acquisition conditions, facilitating the model's learning of skeletal structure commonalities rather than specifics of the acquisition conditions. [7]

**- Image enhancement**: Applying techniques such as rotation, flipping, scaling, and potentially sharpening (as X-ray images are primarily grayscale, making it easier to enhance images through sharpening), to increase data diversity, thereby enhancing the model's adaptability and robustness to new data. Furthermore, to further enhance the model's generalization ability, more advanced image enhancement techniques such as random cropping, adding Gaussian noise, and using GANs (Generative Adversarial Networks) to generate synthetic images were introduced. These methods can simulate different imaging conditions and skeletal variations, providing additional challenges and learning opportunities for model training. For the dataset partitioning, we adopted a ratio of 70% of the data for training, 15% for validation to monitor model training progress and adjust parameters, and the final 15% for testing to assess the model's ultimate performance. Ensuring that these three parts represent the diversity of the entire dataset is crucial. Some of the datasets are shown Fig 1.

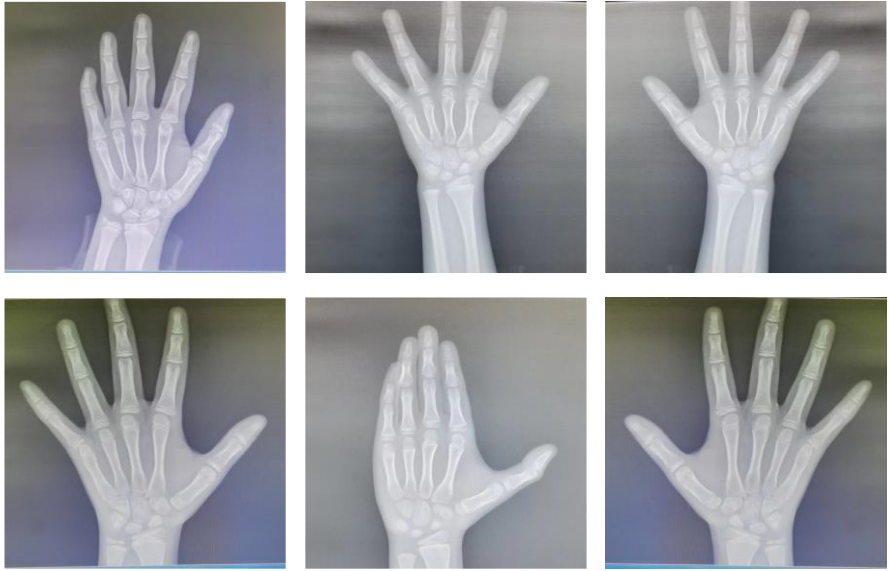

**Fig. 1.** X-ray of bone age of some pediatric patients in Huaibei City Hospital

### 3.2    Model Architecture

The model architecture integrates Convolutional Neural Networks (CNNs) and Transformer technologies. CNNs are responsible for extracting rich local features from X-ray images, while Transformers handle the integration and relational modeling of these features globally.

- **Feature Extraction Layer (CNN)**: We used a pre-trained CNN model (DenseNet) as the feature extractor to derive high-level visual features from the X-ray images. This layer captures the morphology and structural details of the bones, which are critical for accurate bone age prediction. We then employed Xception for fine-tuning training following the pre-trained model to ensure the model's applicability to our dataset.

- **Feature Integration Layer (Transformer)**: The feature maps output by the CNN layer are input into the Transformer model. Utilizing a Fast-Attention Mechanism, the Transformer can capture long-distance dependencies between features, thus offering a comprehensive understanding of the interactions between different parts of the skeleton and the overall impact on bone age. This layer automatically identifies the features most crucial for bone age prediction, thereby improving prediction accuracy.

- **Output Layer**: Finally, the model transforms the Transformer's output into the final bone age prediction value through a fully connected layer. In addition, we introduced Dropout techniques and L2 regularization to reduce the risk of overfitting, ensuring the model's good generalization ability. A schematic diagram of the model is shown in Figure 2.

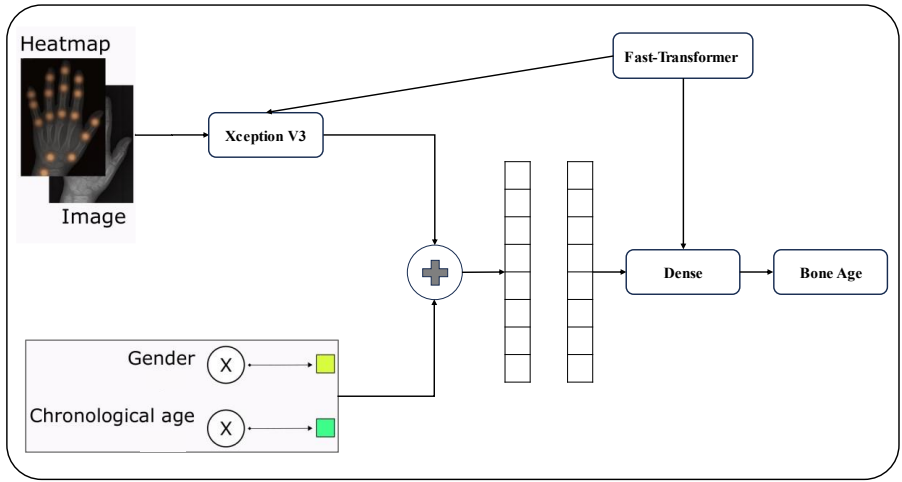

**Fig. 2.** CNN-Transformer Model.

### 3.3 Training and Evaluation Details

The model was trained using a cross-entropy loss function and the Adam optimizer. By monitoring loss and accuracy on the validation set, early stopping was employed to prevent overfitting. Moreover, Mean Squared Error (MSE) was used as the primary evaluation metric to quantify the difference between the model's predicted bone age and the actual bone age. [8]

### 3.4 CBR

Case-based Reasoning (CBR) is a technology for problem-solving that mirrors human learning by referencing past experiences. Unlike traditional methods that rely on universal rules, CBR uses a database of previous cases to inform decision-making for new, similar challenges.[9] It's a part of the Knowledge-based System (KBS) domain, utilizing a knowledge base of individual cases to provide reasoned solutions. Originating from the artificial intelligence field by Schank and Abelson, CBR operates on the principle of comparing new problems to past ones to find and apply similar solutions.[10] The process doesn't adhere to a specific algorithm but follows a general framework that includes case collection, analysis, and application to assist in decision-making. This approach allows for a systematic, logical process tailored to the specifics of each new situation, incorporating additional knowledge as needed for case evaluation and revision. CBR's cyclical process involves four key steps, making it adaptable to a wide array of problems by leveraging historical data and knowledge bases.[11]

**Retrieval**

Upon receiving a new case query, our designed Case-based Reasoning (CBR) system automatically recognizes the patient's X-ray, extracts useful features, and integrates these with the case.[12] It then retrieves similar cases from the case database as references to provide recommendations for the new case to doctors. Here, the CBR system can identify useful features within an X-ray and match these with cases in the database to assist in diagnosis.[13]

**Reuse, Adaptation**

The process of case reuse/adaptation involves finding cases from the past that match the new case. Given the low likelihood of an exact match between new and old cases, new cases are identified through screening and adaptation based on differences. Users can then review similar cases one by one to find the desired outcomes. New knowledge and data are saved in the database for reuse, increasing the accuracy of responses over time.[14] We began by extracting features from 1,233 cases from 2023, storing different case characteristics in the database. We then retrieved and recommended 100 new cases, saving these new cases for future reference.[15]

**Revision, Verification**

After obtaining a result from the previous step, it is sent back for case verification to revise and verify the accuracy of the solution. Automatic revision and verification processes can be applied to specific cases. Given the potential for significant or minimal differences between new and old cases, hospital professionals can make their own adjustments. Cases that significantly differ from previous diagnoses are corrected by a senior doctor with 20 years of experience after the reuse process.[16]

**Retention, Learning**

After a new case is verified, it is added to the case database. Alongside adding new cases, ineffective or incorrect cases must be removed from the database, and tasks to merge or reorganize repetitive or related cases are conducted. This aims to reduce the future volume of cases, prevent search speed impact, and increase case accuracy. This method addresses the challenge of managing a large CBR database. A flowchart of this process is shown in Figure 3.

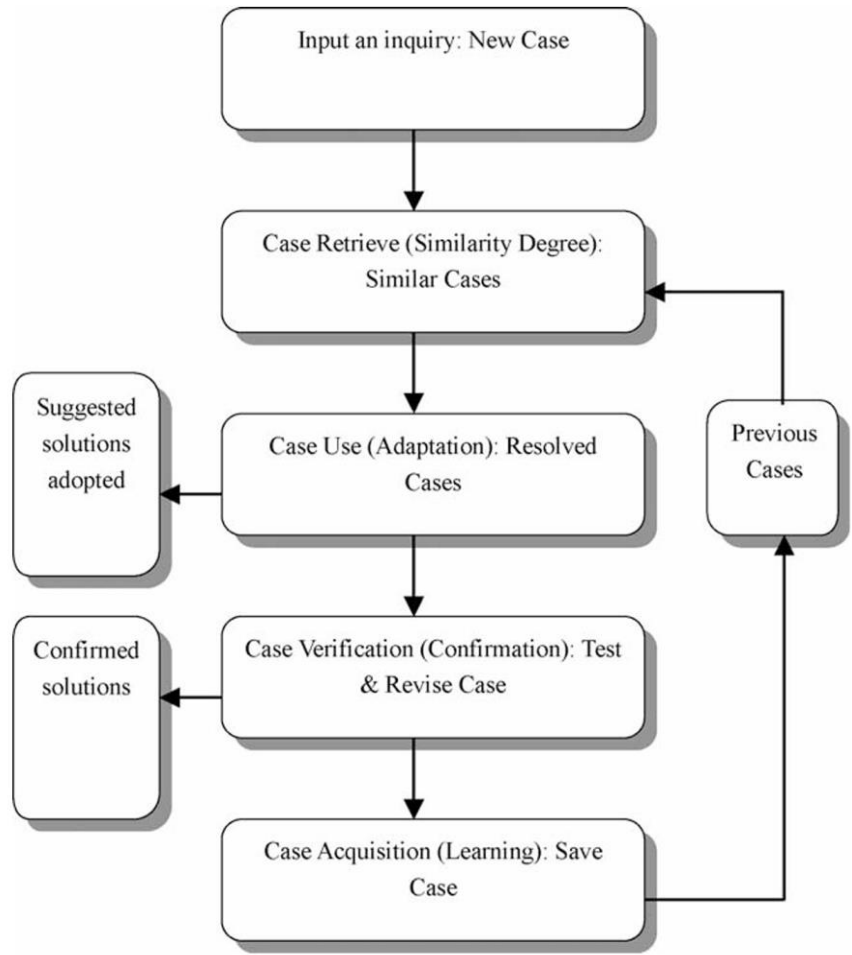

**Fig. 3.** CBR cycle.

### 3.5 Main Method

The aim of this study is to enhance the screening efficiency for children with developmental delays. This research integrates Case-Based Reasoning (CBR) with the use of a screening scale for children aged 0-6 years.[17]
The CBR operational flowchart for this study is as follows:

**1. Input New Case**

Input new cases awaiting screening into the system. Enter the basic information of the case and answer the classification screening questions prompted by the system, such as bone age, language development, social development, motor development, sensory, and cognitive development.[18]

**2. Analyze Inquiry**

Determine whether a new case has developmental delays according to the system's screening procedures.

**3. Determine Weight**

Based on weights provided by medical professionals, the system undergoes internal setup anew. Therefore, this system does not need to assign weight indicators additionally; it can automatically skip the weight setup step and proceed to the next.

**4. Case Selection**

The case base stores many screening pieces of information of children with developmental delays for reference. It offers different suggestions based on the case's conditions.[19]

**5. Similarity Rule**

Compare each attribute of the new case against those in the case base. The case with the highest similarity is given priority. The system will provide 10 sets of similar cases for medical professionals to assess their appropriateness. We employ a CBR similarity algorithm, which is:

$$Similarity\left(f_{input_j}, f_{history_i}\right) = \frac{\sum_{j=1}^{n}\left(W_j \cdot sim\left(f_{input_j}, f_{history_{ij}}\right)\right)}{\sum_{j=1}^{n} W_j}$$

The similarity metric Similarity($f_{input_j}, f_{history_i}$) is quantified through a weighted sum of similarities across multiple indices, where $f_{input_j}$ represents the value of the $j^{th}$ index for the input case, and $f_{history_{ij}}$ represents the value of the $j^{th}$ index for the $j^{th}$ historical case. The function sim($f_{input_j}, f_{history_i}$) computes the similarity for the $j^{th}$ index between the input case and the historical cases. The weight Wj corresponds to the importance of the $j^{th}$ index in determining the overall similarity, indicating that different indices contribute unequally to the similarity score. The sum of these weighted similarities is then normalized by the sum of the weights $\sum_{j=1}^{n} Wj$, to account for the varying number of indices and their weights. This normalization ensures that the similarity score remains consistent and comparable across different cases. The resulting similarity score ranges from 0 to 1, where a higher score indicates a greater degree of similarity between the input and historical cases.[20]

**4. Revise case**

When the solutions derived from the selected similar cases are deemed unsuitable for the current case, medical professionals can apply their expert knowledge to revise these solutions. This adjustment process ensures that the solutions are tailored to the specific circumstances of the case at hand, ultimately leading to the determination of the final screening outcomes.[21]

**5. Save case**

Cases are stored in the case base to enhance its completeness and to reinforce the system's self-learning mechanism. The developmental screening scale for children aged 0–6 comprises six main categories: fundamental physiological examination indicators (12 indicators), language and communication development (31 questions), social interaction (34 questions), gross motor skills (36 questions), fine motor skills (31 questions), and sensory and cognitive development (35 questions). The questions within each category are ordered according to developmental competencies. There are 19 age groups spanning from 0 to 72 months, with several sub-questions per age group. Each sub-question has three response options: Yes, No, and I don't know. An "I Don't Know" response may indicate that the primary caregiver has not observed the behavior in question or that the query is too ambiguous for a clear answer. A high frequency of "I Don't Know" responses suggests the assessment scale may be flawed and should not be utilized.

The "I Don't Know" response serves as a crucial metric to gauge the reliability of the assessment scale. Statistical analysis of "I Don't Know" responses dictates that if more than 10% (16 questions) are marked as such, the results are deemed unreliable. The reasons might be:

1) The respondent, typically the primary caregiver, lacks knowledge on how to observe their child, leading to a generally inaccurate report of developmental status.
2) The respondent may deliberately circumvent questions due to personal reasons or external influences.
3) The respondent might be suffering from psychosis.

Regardless of the cause, when "I Don't Know" exceeds 10%, a diagnostic assessment should follow. To determine developmental delay, calculate the actual age between basal and ceiling levels and divide by the physical age to obtain a ratio. This ratio is then used to assess developmental delay, with each scenario detailed in the Table 1. In this system, the weighting of similarity is based on the infants' and toddlers' physical age and the screening questions across the six main categories, as presented in Table 2.

Through the screening procedures developed in this research, it can be determined whether a child is developmentally delayed. If special circumstances are present, medical professionals can make revisions before the case is saved in the database for future reference.

**Table 1.** Organization of judgmental data of the screening scale

| Criteria | Range | Developmental Status |
|---|---|---|
| Actual age/physical age | >0.75 | Developmental delay |
| | <0.70 | Normal development |
| | 0.7–0.75 | At the edge of normal |
| Peak level – basal level | ≥6.00 | Developmental status is too wide |
| | <6.00 | Developmental status is normal |

Table 2. Weight distributions of the screening scale

| Items in comparison | Level | Weight of sim degree |
|---|---|---|
| Actual age | | 20 |
| Language and communication development | Basal level | 8 |
| | Peak level | 8 |
| Social personality development | Basal level | 8 |
| | Peak level | 8 |
| Rough movement/motor development | Basal level | 8 |
| | Peak level | 8 |
| Delicate movement/motor development | Basal level | 8 |
| | Peak level | 8 |
| Sensory and cognitive development | Basal level | 8 |
| | Peak level | 8 |

## 4　Results

In order to verify the feasibility of the diagnostic screening system for children with developmental delay, this study employed real screening cases. The information stored in the system database includes cases of children with developmental delay selected by a screening center in 2023, totaling 100 cases. In addition, cases from 2022 were used for system verification. Fifty data were randomly selected for the case-based reasoning diagnostic system to perform diagnosis.

The case verification of the system selected five previously searched cases as subjects to calculate the system's reasoning accuracy. The overall system verification results, as shown in Table 3, indicate that the average similarity is 0.9217, and the system's average accuracy rate is 0.9129. These two figures demonstrate a high level of average accuracy and similarity for the old cases reasoned by the system. Therefore, it shows a high feasibility of this diagnostic screening system, which can serve as a support system in decision-making for the staff of the screening center while diagnosing children with developmental delay.

Table 3. Overall result of system verification

| Ranking | Average Similarity | Standard Deviation (Similarity) | Average Accuracy | Standard Deviation (Accuracy) |
|---|---|---|---|---|
| 1 | 0.9531 | 0.0516 | 0.9256 | 0.1054 |
| 2 | 0.9355 | 0.0564 | 0.9123 | 0.1046 |
| 3 | 0.9167 | 0.0581 | 0.9132 | 0.1028 |
| 4 | 0.9048 | 0.0602 | 0.9024 | 0.1033 |
| 5 | 0.8982 | 0.0627 | 0.9112 | 0.1019 |
| Mean | 0.9217 | 0.0578 | 0.9129 | 0.1036 |

## 5    Conclusions

The theme of this research is the use of Case-Based Reasoning (CBR) to enhance the screening efficiency for children with developmental delays, aiming to introduce a tailored screening system for this demographic. Such screening needs to be rapid, effective, and cost-efficient, diverging from conventional assessments by focusing on identifying the causes of developmental delays, evaluating the developmental status, and the severity thereof. This process encompasses professionals across all relevant fields. To prevent the depletion of societal resources by screening costs, the operational aspect of screening can be utilized to filter children showing signs of developmental delays before proceeding to further evaluations and early intervention assistance.

Furthermore, the screening system designed in this study aids in identifying children with developmental delays. It is capable of automatically generating screening charts, developmental statuses, ranges, and judgment results based on case specifics. Such functionality saves valuable time that medical professionals would otherwise spend making manual assessments. The CBR system selects the five most similar cases from an existing case database as references for the users. Medical professionals can then consult these precedent cases to inform their evaluations of new cases. Therefore, this system not only significantly improves screening efficiency but also provides medical professionals with a convenient set of tools for the screening of children with developmental delays.